\documentclass[letterpaper]{article} 
\usepackage{aaai2027}
\usepackage[hyphens]{url}  
\usepackage{graphicx} 
\usepackage{natbib}  
\usepackage{caption} 
\usepackage{amsmath}
\usepackage{amssymb}
\usepackage{amsfonts}

\usepackage{amsthm}

\usepackage{amsthm}

\usepackage{algorithm}
\usepackage{algorithmic}

\usepackage{booktabs}
\usepackage{newfloat}
\usepackage{listings}
\DeclareCaptionStyle{ruled}{labelfont=normalfont,labelsep=colon,strut=off} 
\floatstyle{ruled}
\newfloat{listing}{tb}{lst}{}
\floatname{listing}{Listing}

\usepackage{multirow}
\usepackage{colortbl}
\usepackage{booktabs}
\usepackage{array}

\AtBeginDocument{%
  }
\usepackage{booktabs}

\title{NANQ: Noise-Floor-Aware Mixed-Precision Non-Uniform Quantization for Analog Compute-in-Memory}

\author{
    Written by AAAI Press Staff\textsuperscript{\rm 1}\thanks{With help from the AAAI Publications Committee.}\\
    AAAI Style Contributions by Peter Patel Schneider,
    Sunil Issar,\\
    J. Scott Penberthy,
    George Ferguson,
    Hans Guesgen,
    Francisco Cruz\equalcontrib\corresponding,
    Marc Pujol-Gonzalez\equalcontrib\corresponding
}
\affiliations{
    \textsuperscript{\rm 1}Association for the Advancement of Artificial Intelligence\\

    1101 Pennsylvania Ave, NW Suite 300\\
    Washington, DC 20004 USA\\
    proceedings-questions@aaai.org
}

\author{
Yizhe Chen\textsuperscript{\rm 1}\equalcontrib,
Wenshuai Yao\textsuperscript{\rm 2}\equalcontrib,
Saiya Wang\textsuperscript{\rm 1},
Yuannuo Feng\textsuperscript{\rm 1},
Wenbo Qi\textsuperscript{\rm 3},\\
Kechao Tang\textsuperscript{\rm 2},
Ngai Wong\textsuperscript{\rm 3},
Wenyong Zhou\textsuperscript{\rm 3}\corresponding,
Wang Kang\textsuperscript{\rm 1}\corresponding
}

\affiliations{
\textsuperscript{\rm 1}School of Integrated Circuit Science and Engineering, Beihang University, Beijing, China\\
\textsuperscript{\rm 2}School of Integrated Circuits, Peking University, Beijing, China\\
\textsuperscript{\rm 3}Department of Electrical and Computer Engineering, The University of Hong Kong, Hong Kong SAR, China
}

\begin{document}

\maketitle

\begin{abstract}
Analog compute-in-memory (CIM) enables energy-efficient neural network inference, but device variation and read noise can severely degrade low-bit quantized models. Existing CIM-oriented quantization methods mainly minimize ideal quantization error, ignoring the hardware noise floor and thus causing inefficient precision allocation.
We propose NANQ, a noise-aware mixed-precision non-uniform quantization framework for analog CIM. NANQ models magnitude-dependent weight noise from measured responses of an eFlash CIM array and converts the noise profile into an adaptive quantization density, assigning finer resolution to low-noise regions while avoiding ineffective precision in noise-dominated regions. It further assigns layer-wise bit-widths by identifying each layer's precision saturation point under hardware noise using a unified threshold.
On-chip experiments on an eFlash CIM SoC show that, under 2-bit
weight-magnitude quantization, NANQ improves vision-model accuracy by
$8.05$ percentage points and reduces language-model PPL by $54.7\%$ on
average over PowerQuant. Mixed-precision NANQ captures most of the gains
obtainable from additional quantization resources with only $3.2$--$3.8$
equivalent bits.
\end{abstract}


\section{Introduction}

The exponential growth in deep neural network (DNN) complexity has driven the need for energy-efficient inference accelerators~\cite{11132743}. Analog compute-in-memory (CIM) architectures have emerged as a promising solution, leveraging in-situ computation within memory arrays to achieve significant improvements in energy efficiency and computational density compared to traditional digital processors~\cite{Mao2025HyIMC, chen2025reconfigurable}. By storing neural network weights directly in memory cells and performing matrix--vector multiplications through analog operations, CIM systems can reduce data movement overhead and enable massively parallel computation~\cite{Guo2017Fast}.

However, analog CIM implementations face fundamental challenges that limit their practical deployment~\cite{11219045}. Device variations, thermal noise, and nonlinearities inherent in analog circuits introduce computational errors that can severely degrade neural network accuracy~\cite{8806938}. These noise sources create discrepancies between the expected weights and the weights programmed in CIM arrays~\cite{10363786, 8702401}.
\begin{figure}[!t]
\centering
\includegraphics[width=1.0\columnwidth]{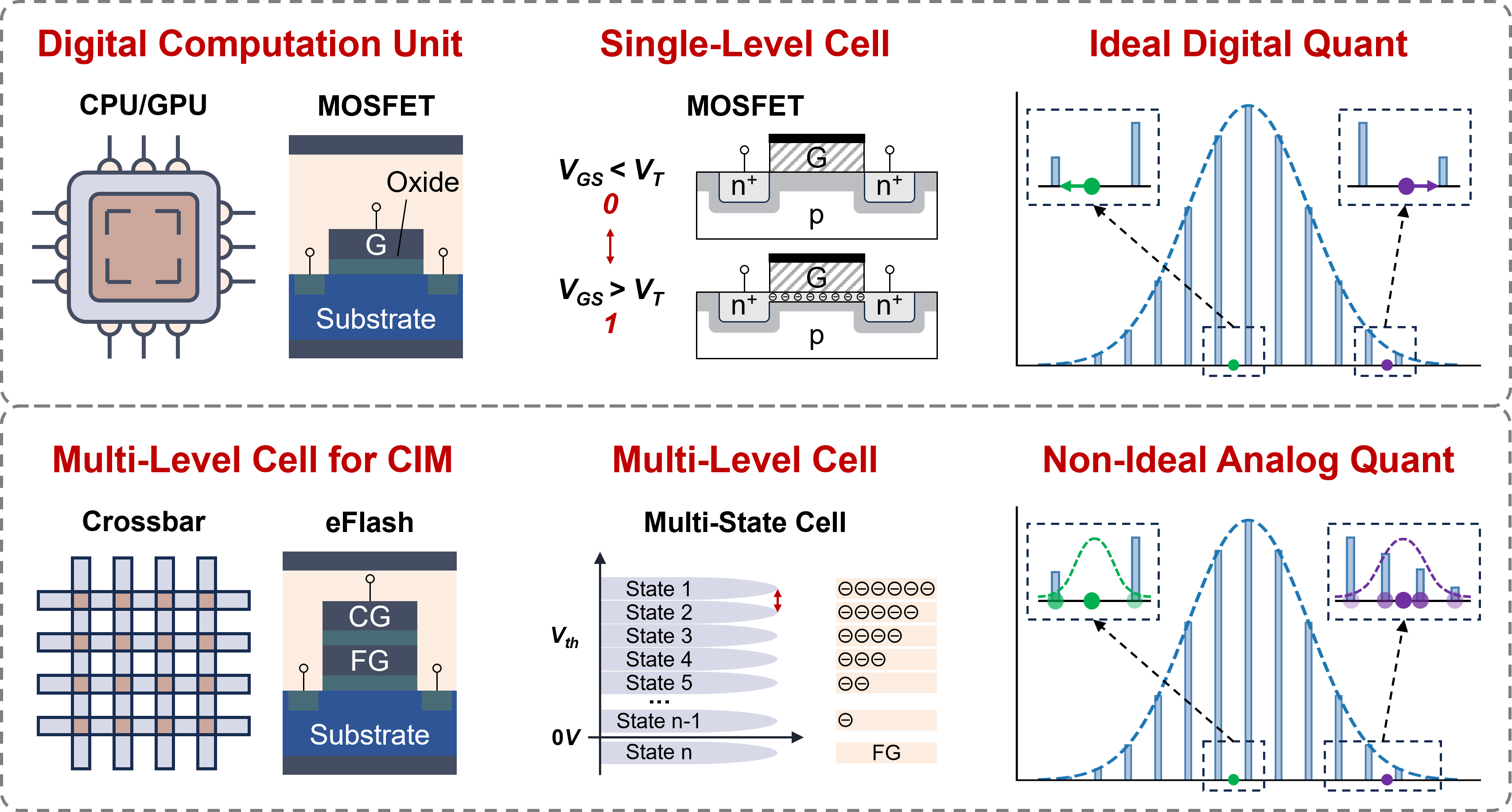}
\caption{Fixed quantization levels spread into noisy distributions on analog hardware, causing diminishing returns with higher precision.}
\label{fig:Intro}
\vspace{-0.2cm}
\end{figure}

Due to the discrete nature of memory cell states and the limited precision of analog sensing circuits, CIM architectures inherently require model quantization~\cite{bai2023cimq, sun2024cim2pq, chen2024device}. Previous CIM-oriented quantization schemes have fully explored the potential of uniform quantization, including mixed-precision quantization and extreme binary or ternary quantization, but have less explored non-uniform quantization, which is ideal for handling the distribution shifts caused by hardware non-idealities. 

As shown in Figure~\ref{fig:Intro}, during the ideal digital quantization process, continuous weight values are mapped to the nearest discrete quantization levels on either side. However, after being programmed on analog CIM, each quantization level broadens into a noisy distribution whose width depends on the hardware noise strength.

To address these challenges, we propose noise-aware non-uniform quantization
(NANQ), a hardware-aware framework for robust DNN inference on analog CIM
accelerators. NANQ jointly considers noise variation across weight regions
and network layers to optimize intra-layer quantization levels and inter-layer
bit-widths. The main contributions are summarized as follows:

\begin{itemize}
    \item We analyze the interaction between quantization error and chip-measured magnitude-dependent weight noise in analog CIM, showing that the benefit of increasing bit-width saturates once quantization error falls below the hardware noise floor.
    \item We propose NANQ, a training-free non-uniform quantization method that converts a hardware noise profile into an adaptive quantization density and constructs quantization boundaries by cumulative-density partitioning.
    \item We introduce a noise-aware mixed-precision allocation strategy that determines each layer's saturation bit-width from hardware-aware loss curves using a unified threshold, avoiding redundant precision without retraining or combinatorial search.
\end{itemize}

On-chip evaluations across CNNs, ViTs, and language models show that NANQ
achieves the best quantized result in 38 of 45 model--bit-width
configurations, with particularly consistent gains at low precision. Under
matched quantization resources, mixed-precision NANQ further outperforms all
evaluated baselines across the tested models.

\section{Related Work}

\begin{figure}[!t]
\centering
\includegraphics[width=1.0\columnwidth]{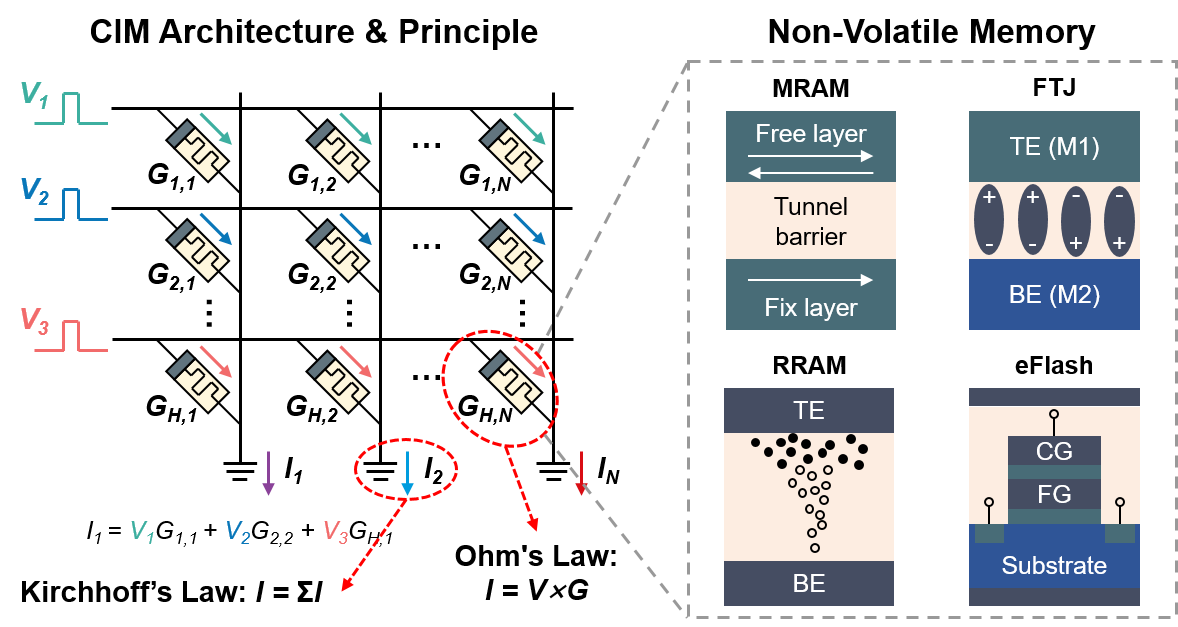}
\caption{Schematic of an analog compute-in-memory array based on nonvolatile memory (NVM-CIM).}
\label{fig:cim}
\end{figure}

\paragraph{Analog CIM and Device Non-Idealities}

As illustrated in Figure~\ref{fig:cim}, analog CIM maps neural network weights
to device conductance states and converts input activations into word-line
voltages. The resulting cell currents are accumulated along each column,
enabling parallel matrix--vector multiplication in the analog domain. 
This
computing principle can be implemented using magnetic, resistive, and
floating-gate memory devices.

Practical devices, however, deviate from ideal linear computation. Process
variation, stochastic programming, oxide defects, and charge trapping
introduce quasi-static cell-to-cell conductance variation that persists across
inference operations~\cite{cellvar}. Nonlinear weight encoding, input-voltage
conversion, and device \(I\)--\(V\) characteristics further produce systematic
input- and weight-dependent distortions, which are repeatable under fixed
operating conditions and can be partially mitigated through
calibration~\cite{nonlinear}. In addition, thermal, flicker, and peripheral
readout noise cause temporal fluctuations that vary across inference
passes~\cite{nuo}. Together, these non-idealities cause the accumulated
column outputs to deviate from ideal matrix--vector multiplication.

\paragraph{Non-Uniform Quantization}


Non-uniform quantization improves low-bit representation by redistributing
quantization levels over the weight range. APoT constructs structured levels
using sums of powers of two to better represent long-tailed weight
distributions~\cite{li2019additive}, while PowerQuant searches layer-wise power
transformations for data-free non-uniform quantization~\cite{yvinec2023powerquant}.
SqueezeLLM further uses second-order sensitivity to construct non-uniform
codebooks and preserves sensitive outliers with sparse high-precision
representations~\cite{kim2023squeezellm}. Nevertheless, these methods primarily
optimize weight distributions, digital reconstruction error, or model
sensitivity, without considering the magnitude-dependent hardware noise of
analog CIM.

\paragraph{Mixed-Precision Quantization}

Mixed-precision quantization assigns different bit-widths to network layers
to better balance model accuracy and resource consumption. PNMQ jointly
optimizes parametric non-uniform quantization grids and layer-wise bit-widths
under a specified compression budget without retraining~\cite{mixed}. OMPQ
uses network orthogonality as an efficient proxy for layer importance and
determines the bit-width configuration through linear programming
\cite{ma2023ompq}. More recently, InfoQ measures the global impact of layer-wise
quantization on information flow and formulates precision allocation as an
integer linear programming problem~\cite{akbulut2026infoq}. Although these methods
efficiently capture layer-dependent precision requirements, their allocation
criteria are derived from ideal digital quantization and do not account for
the hardware noise floor of analog CIM. 


\section{Methodology}

\begin{figure}[!t]
\centering
\includegraphics[width=1.0\columnwidth]{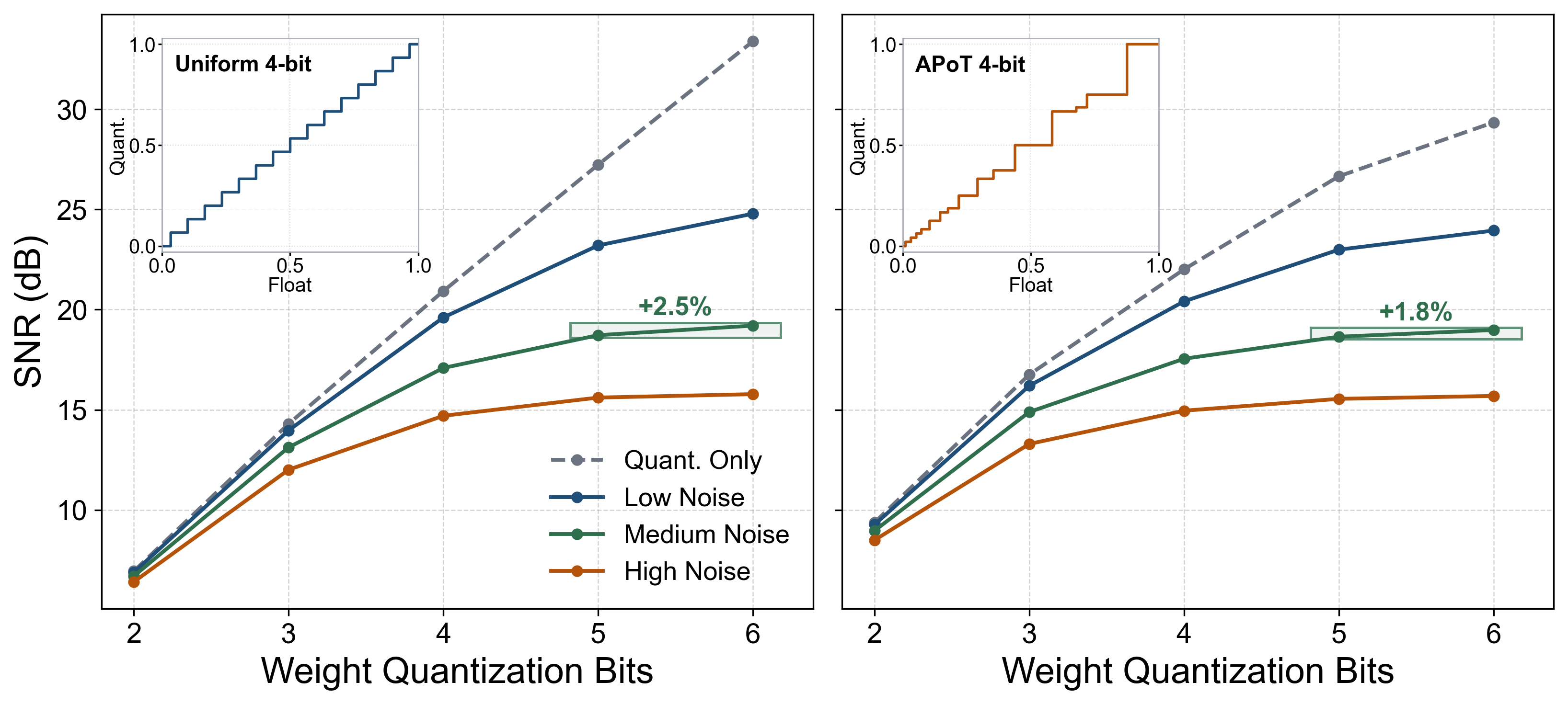}
\caption{Output SNR versus weight quantization bit-width under different noise conditions.}
\label{fig:motivation}
\end{figure}

\subsection{Motivation}

As shown in Figure~\ref{fig:motivation}, conventional quantization determines
quantization levels primarily by minimizing ideal weight-domain reconstruction
error. After the quantized weights are mapped to physical conductance states,
however, each level is further perturbed by magnitude-dependent hardware noise.
Once the local quantization error falls below the hardware noise floor, finer
quantization provides little effective improvement. Consequently, quantization
boundaries derived solely from ideal reconstruction error may waste resolution
in noise-dominated regions while providing insufficient resolution in more
reliable regions. This mismatch motivates adaptively redistributing intra-layer
quantization levels according to the measured hardware noise profile.

\begin{figure}[!t]
\centering
\includegraphics[width=1.0\columnwidth]{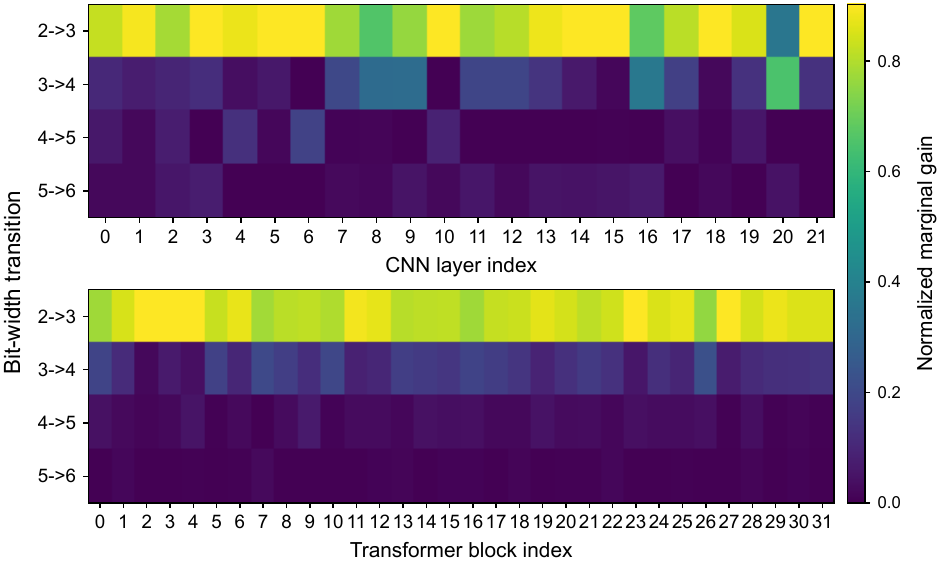}
\caption{Layer-wise marginal gains from increasing weight precision under analog CIM noise: ResNet-20 on CIFAR-100~\cite{cifar10} (top) and Llama-3.2-3B on WikiText-2~\cite{wikitext} (bottom).}
\label{fig:layer}
\end{figure}

We further examine whether different layers benefit equally from increasing
weight precision. Using uniform quantization, we evaluate each layer
independently under hardware noise. Let $\mathcal{L}_l(b)$ denote the validation
loss when layer $l$ uses $b$-bit weight precision, and let $N_l$ denote its
number of weights. Since increasing the precision of this layer by one bit
introduces $N_l$ additional weight bits, we define the marginal loss reduction
per added weight bit as
\begin{equation}
G_l(b)=
\frac{\max\!\left(\mathcal{L}_l(b)-\mathcal{L}_l(b+1),0\right)}
     {N_l}.
\end{equation}
As shown in Figure~\ref{fig:layer}, different layers exhibit substantially
different marginal gains at the same bit-width, while the gain of each layer
generally decreases as precision increases. Some layers approach saturation at
low precision, whereas others continue to benefit from additional bits. This
heterogeneity is consistently observed in both CNNs and LLMs, indicating that
a uniform bit-width may allocate redundant precision to already saturated
layers. This observation motivates identifying a layer-specific saturation bit-width
and avoiding redundant precision allocation to layers whose hardware-aware
gains have already diminished.

\subsection{Theoretical Analysis}

We analyze why increasing the quantization bit-width provides diminishing
returns under analog hardware noise. For a weight region around $w$, the
total error variance can be approximated as
\begin{equation}
\sigma_{\mathrm{tot}}^2(b,w)
=
\sigma_q^2(b,w)+\sigma_h^2(w),
\end{equation}
where $\sigma_q^2(b,w)$ and $\sigma_h^2(w)$ denote the local quantization-error
variance and hardware noise variance, respectively. Assuming a local
quantization step $\Delta_b(w)$ at bit-width $b$, the quantization-error
variance can be approximated as
\begin{equation}
\sigma_q^2(b,w)
\approx
\frac{\Delta_b^2(w)}{12}
\propto 2^{-2b}.
\end{equation}
The corresponding output SNR is therefore expressed as
\begin{equation}
\mathrm{SNR}(b,w)
=
\frac{P_{\mathrm{sig}}(w)}
{\sigma_q^2(b,w)+\sigma_h^2(w)},
\end{equation}
where $P_{\mathrm{sig}}(w)$ denotes the signal power associated with the
considered weight region.

Increasing $b$ exponentially reduces $\sigma_q^2(b,w)$ but does not reduce
the hardware noise term. Once
$\sigma_q^2(b,w)\ll\sigma_h^2(w)$, the SNR approaches
$P_{\mathrm{sig}}(w)/\sigma_h^2(w)$, and additional bits provide little
improvement. Moreover, weight regions or network layers with stronger
hardware noise reach this precision-saturation point at lower bit-widths.
Therefore, quantization resolution should be concentrated in relatively
reliable weight regions, while additional bits should be assigned only to
layers that still exhibit meaningful marginal gains.

This analysis establishes the precision-saturation effect but does not
uniquely determine the optimal quantization density. Motivated by this
observation, we adopt a lightweight inverse-noise density in the following
section.

\begin{figure*}[]
\centering
\includegraphics[width=1.0\textwidth]{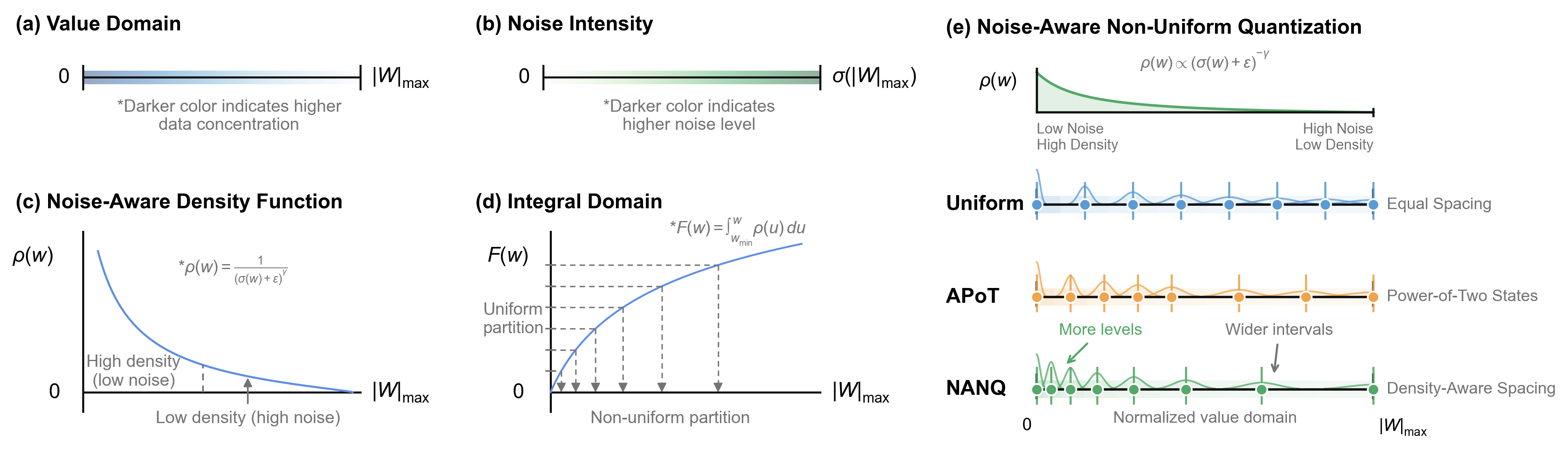}
\caption{Overview of our NANQ method. Building on existing quantization methods, NANQ adjusts quantization points based on the noise sensitivity metric of values prior to quantization.}
\label{fig:nanq}
\end{figure*}

\subsection{Adaptive Quantization Algorithm}

As illustrated in Figure~\ref{fig:nanq}, NANQ converts the
magnitude-dependent hardware noise profile into a non-uniform quantization
density, which is then used to construct noise-adaptive quantization
boundaries.

Figure~\ref{fig:nanq}(a) and (b) illustrate the normalized weight-magnitude
domain and its corresponding noise-intensity profile, respectively. We
characterize this profile using a noise strength function $\sigma(w)$, where
$\sigma(w)\geq0$ represents the relative noise intensity at magnitude $w$.
For the evaluated eFlash-based CIM SoC, $\sigma(w)$ is obtained directly from
repeated on-chip readback measurements at different programmed weight
magnitudes. Here, $\sigma(w)$ provides a device-characterization prior for
constructing the quantizer rather than serving as an explicit inference-noise
simulator. Quantizer calibration and end-to-end evaluation are performed using
real-chip responses. This formulation does not impose a specific analytical
form on the underlying noise model and can therefore accommodate different
analog CIM devices and technologies. Based on $\sigma(w)$, we define the inverse
quantization-density function shown in Figure~\ref{fig:nanq}(c):
\begin{equation}
\rho(w)=\frac{1}{\left(\sigma(w)+\epsilon\right)^\gamma},
\end{equation}
where $\gamma\geq0$ controls the adaptation strength and $\epsilon>0$
ensures numerical stability. Through this inverse relationship, low-noise
regions receive a higher quantization density, whereas fewer quantization
resources are assigned to regions dominated by hardware noise.

The parameters $\gamma$ and $\epsilon$ provide flexible control over the
quantization-level distribution. When $\gamma=0$, the density becomes
constant, and the method reduces to uniform quantization regardless of the
noise characteristics. As $\gamma$ increases, the density becomes more
sensitive to noise variation, strengthening the redistribution between
high-noise and low-noise regions. Meanwhile, $\epsilon$ prevents numerical
instability and excessive concentration of quantization levels in regions
where $\sigma(w)$ approaches zero.

To transform the continuous density into discrete quantization intervals,
we compute the cumulative density shown in Figure~\ref{fig:nanq}(d):
\begin{equation}
F(w)=\int_{0}^{w}\rho(u)\,du.
\end{equation}
Given $Q$ magnitude intervals determined by the selected magnitude bit-width,
the boundaries $\{e_i\}$ are obtained by equally partitioning the cumulative
density:
\begin{equation}
F(e_i)=\frac{i}{Q}F(w_{\max}),
\qquad i=0,1,\ldots,Q.
\end{equation}
The reconstruction level and signed quantized weight are then defined as
\begin{equation}
q_i=\frac{e_i+e_{i+1}}{2},
\qquad
\widehat{w}=\operatorname{sign}(w)q_i,
\quad |w|\in[e_i,e_{i+1}).
\end{equation}
The empirical weight distribution in Figure~\ref{fig:nanq}(a) is shown only
to visualize parameter concentration and is not used to construct
$\rho(w)$. Since each interval contains the same cumulative noise-aware
density, low-noise regions receive narrower intervals and finer resolution,
whereas high-noise regions are represented using wider intervals.

As summarized in Figure~\ref{fig:nanq}(e), uniform quantization uses equally
spaced levels, while APoT remains constrained by predefined power-of-two
combinations. In contrast, NANQ directly converts the hardware noise profile
into density-aware quantization spacing. This CDF-based construction preserves
the fixed budget of $Q$ intervals and generates noise-adaptive boundaries
without explicit iterative boundary optimization.

\subsection{Noise-Aware Mixed-Precision Allocation}

Motivated by the layer-wise differences in precision saturation under hardware
noise, we use a unified saturation threshold to independently determine the
weight precision of each layer. Consider a network with $L$ quantized layers
and a candidate magnitude-bit set
$\mathcal{B}=\{b_{\min},\ldots,b_{\max}\}$, where the sign bit is excluded.
For layer $l$ and candidate bit-width $b$, we apply NANQ quantization and
hardware noise only to this layer while keeping all other layers in full
precision. The resulting noisy quantized weight is denoted by
\begin{equation}
\widehat{W}_l^{b,\gamma,\xi}
=
\widetilde{Q}_{\mathrm{NANQ}}(W_l;b,\gamma,\xi),
\end{equation}
where $\gamma\in\Gamma$ controls the adaptation of the quantization levels to
the hardware-noise profile, and $\xi$ is sampled from the measured noise
distribution $p_{\mathrm{hw}}$. We then define the minimum hardware-aware loss
of layer $l$ at bit-width $b$ as
\begin{equation}
\begin{aligned}
E_l(b)
=
\min_{\gamma\in\Gamma}\;
\mathbb{E}_{\xi\sim p_{\mathrm{hw}}}
\Big[
\mathcal{L}\big(
f(W_{-l},\widehat{W}_l^{b,\gamma,\xi});
\mathcal{D}_{\mathrm{cal}}
\big)
\Big],
\end{aligned}
\end{equation}
where $\mathcal{D}_{\mathrm{cal}}$ denotes the calibration set and $W_{-l}$
represents the full-precision weights of all layers except layer $l$. The
corresponding optimal noise-adaptation parameter is denoted by
$\gamma_l^{*}(b)$.

Finite calibration data and stochastic noise sampling may introduce local
fluctuations into $\{E_l(b)\}_{b\in\mathcal{B}}$. We therefore fit a
monotonically non-increasing loss curve, denoted by $\widetilde{E}_l(b)$.
The fraction of recoverable loss reduction remaining beyond bit-width $b$ is
then defined as
\begin{equation}
R_l(b)=
\frac{
\widetilde{E}_l(b)-\widetilde{E}_l(b_{\max})
}{
\widetilde{E}_l(b_{\min})-\widetilde{E}_l(b_{\max})+\epsilon
}.
\end{equation}
This normalization removes the scale differences among layers and measures the
relative benefit that remains when increasing the precision from $b$ to
$b_{\max}$. Given a global saturation threshold $\tau$, the bit-width of layer
$l$ is selected as
\begin{equation}
b_l^{*}(\tau)=
\min\left\{
b\in\mathcal{B}\mid R_l(b)\leq\tau
\right\}.
\end{equation}
For example, $\tau=0.1$ indicates that the selected bit-width has already
captured at least $90\%$ of the recoverable loss reduction between
$b_{\min}$ and $b_{\max}$. If the total recoverable reduction of a layer is
negligible, we directly set $b_l^{*}=b_{\min}$ to avoid assigning redundant
precision to an already saturated layer.

The final configuration assigns each layer $l$ the pair
$\bigl(b_l^{*}(\tau),\gamma_l^{*}(b_l^{*}(\tau))\bigr)$, where a smaller
$\tau$ generally results in higher precision. To account for cross-layer
error accumulation, we evaluate a small set of candidate thresholds on the
full model and select the point beyond which further decreasing $\tau$
provides negligible end-to-end improvement. The resulting resource cost is
reported using the equivalent average bit-width
\begin{equation}
B_{\mathrm{eq}}(\tau)=
\frac{\sum_{l=1}^{L}N_l b_l^{*}(\tau)}
     {\sum_{l=1}^{L}N_l},
\end{equation}
where $N_l$ is the number of weights in layer $l$.
This procedure requires only layer--bit evaluations and a few full-model
validations, without retraining or combinatorial bit-width search.

\section{Experiments}
\label{sec:experiments}

\subsection{Experiment Setup}
\begin{figure}[!t]
\centering
\includegraphics[width=0.9\columnwidth]{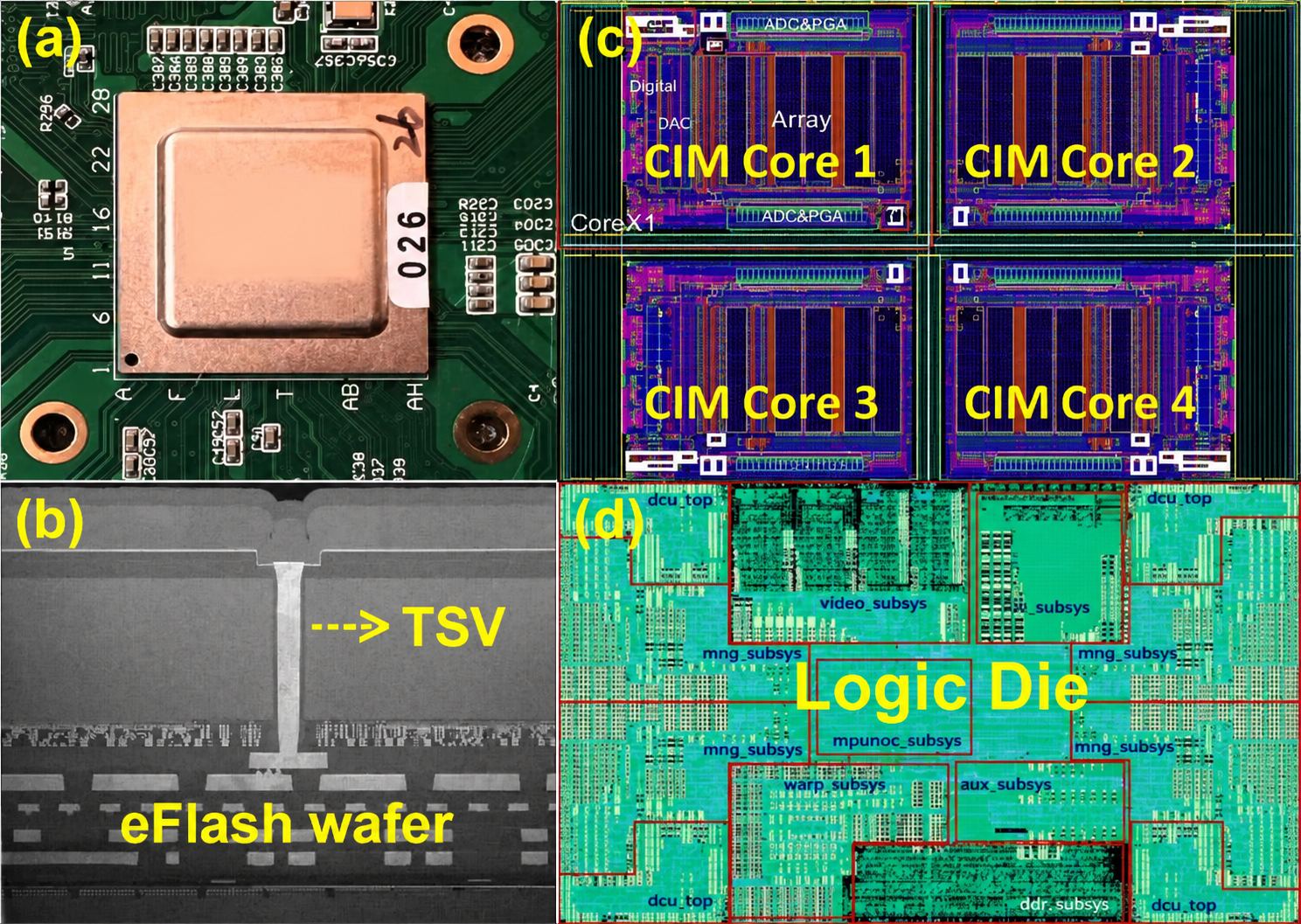}
\caption{Photographs of the 3D eFlash CIM SoC. (a) Packaged SoC and test board; (b) TEM cross-section of the 3D stack; (c) Flash die layout; (d) Logic die layout.}
\label{fig:chip}
\end{figure}

We conduct end-to-end evaluations on representative CNN and Transformer
models using the eFlash CIM SoC shown in Figure~\ref{fig:chip}. For CNNs,
convolutional and fully connected layers are mapped onto the physical CIM
array, while QKV projection and FFN layers are mapped for ViTs and LLMs.
Layers exceeding the on-chip capacity are partitioned into tiles and executed
sequentially, whereas all remaining operations, including batch normalization
and softmax, are performed on an external NPU. Weights are quantized per
channel, while activations are quantized per token.

\subsection{Experiment Results}

\subsubsection{MAC-Level Quantization Analysis}

\begin{figure}[!t]
\centering
\includegraphics[width=1.0\columnwidth]{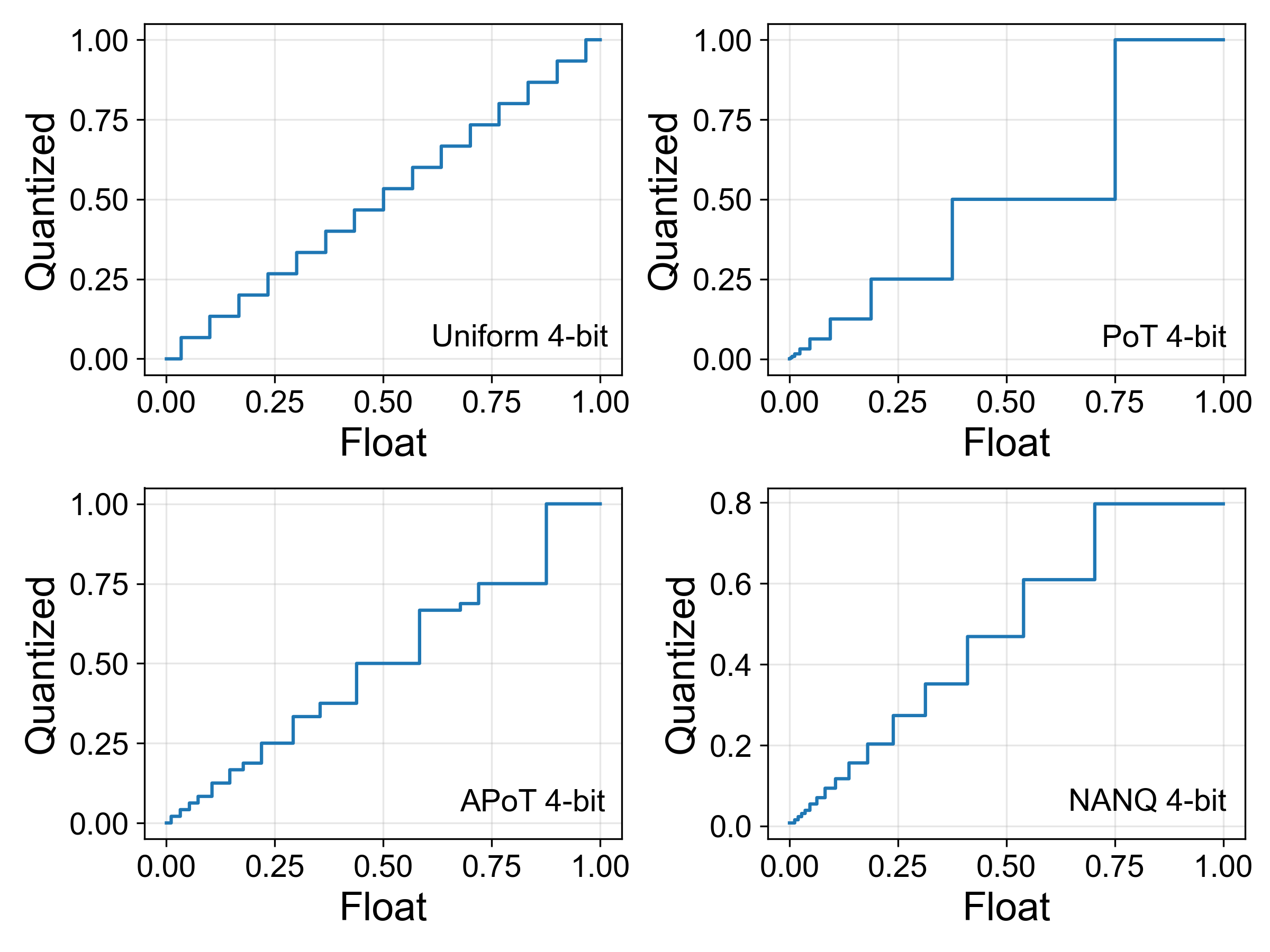}
\caption{Comparison of quantization steps across different methods.}
\label{fig:step}
\end{figure}

As shown in Figure~\ref{fig:step}, uniform quantization distributes levels evenly across the entire range, which does not match the bell-shaped weight distribution and may waste levels in low-density regions. PoT~\cite{zhou2017incremental} concentrates quantization levels near zero, but increasing the bit-width mainly adds levels to the small-magnitude region, leaving limited resolution for larger values. APoT improves the coverage by combining multiple power-of-two terms, yet its level distribution is still restricted by predefined structures. In contrast, NANQ introduces only one hyperparameter to flexibly reshape the quantization levels according to the magnitude-dependent hardware noise profile. It remains training-free and requires only a few inference trials for parameter search, making it more suitable for noisy analog CIM inference.

\begin{figure}[!t]
\centering
\includegraphics[width=1.0\columnwidth]{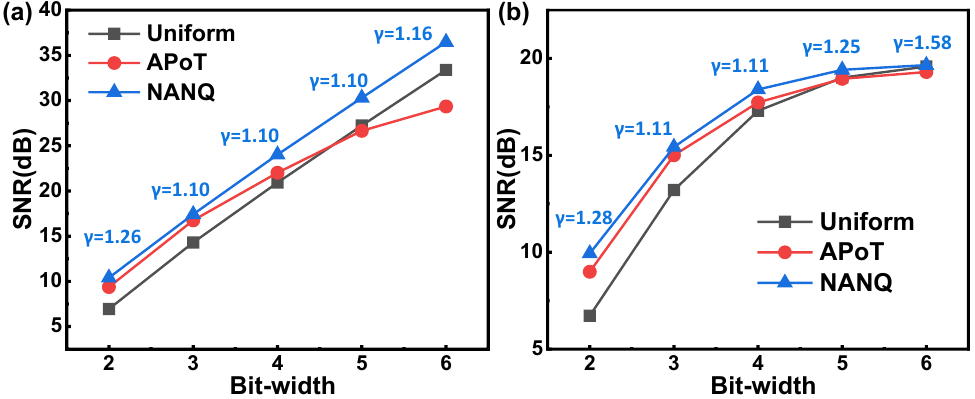}
\caption{Single-MAC output SNR under (a) quantization only and (b) on-chip quantization with hardware noise.}
\label{fig:mac}
\end{figure}

As shown in Figure~\ref{fig:mac}, NANQ achieves higher single-MAC output SNR, particularly under low-bit on-chip inference. Since neural network weights are typically concentrated near zero and the measured on-chip noise profile increases approximately with $|W|$, the small-magnitude region simultaneously exhibits a higher weight density and a lower noise level. NANQ therefore allocates more quantization levels to this region. Under the quantization-only setting, such allocation utilizes the limited quantization levels more effectively and reduces the dominant quantization error. With on-chip hardware noise, NANQ further avoids ineffective precision allocation to large-magnitude regions with stronger noise, leading to a more pronounced robustness advantage at low bit-widths.

\begin{table*}[t]
\centering

\begingroup
\newcommand{\best}[1]{\cellcolor{gray!15}\textbf{#1}}
\setlength{\tabcolsep}{3.5pt}
\renewcommand{\arraystretch}{0.88}
\footnotesize

\textbf{(a) Vision Models: Top-1 Accuracy (\%) $\uparrow$}

\vspace{1.5pt}

\begin{tabular*}{\textwidth}{
    @{\extracolsep{\fill}}
    lll*{6}{c}
    @{}
}
\toprule
Family
& Model
& Method
& Baseline
& $B_m=2$
& $B_m=3$
& $B_m=4$
& $B_m=5$
& $B_m=6$ \\
\midrule

\multirow{4}{*}{\shortstack{ResNet\\[-1pt]\scriptsize CIFAR-100}}
& \multirow{4}{*}{20}
& Uniform
& \multirow{4}{*}{62.89}
& 2.35 & 24.64 & 54.90 & 56.43 & 57.29 \\

& & APoT~\cite{li2019additive}
& {}
& 4.21 & 46.27 & 55.57 & 55.81 & 56.95 \\

& & PowerQuant~\cite{yvinec2023powerquant}
& {}
& 12.63 & 50.84 & 56.61 & \best{58.14} & 57.23 \\

& & NANQ
& {}
& \best{19.84} & \best{53.92} & \best{57.24}
& 57.89 & \best{57.56} \\

\midrule

\multirow{4}{*}{\shortstack{VGG\\[-1pt]\scriptsize CIFAR-100}}
& \multirow{4}{*}{11}
& Uniform
& \multirow{4}{*}{66.88}
& 1.16 & 50.46 & 64.79 & 65.88 & 66.18 \\

& & APoT~\cite{li2019additive}
& {}
& 3.70 & 61.59 & 66.28 & 66.04 & 66.13 \\

& & PowerQuant~\cite{yvinec2023powerquant}
& {}
& 31.86 & 64.33 & 66.40 & 66.22 & \best{66.35} \\

& & NANQ
& {}
& \best{56.27} & \best{66.72} & \best{66.51}
& \best{66.45} & 66.27 \\

\midrule

\multirow{8}{*}{\shortstack{ViT\\[-1pt]\scriptsize ImageNet-1K}}
& \multirow{4}{*}{Tiny}
& Uniform
& \multirow{4}{*}{75.46}
& 3.78 & 61.50 & 70.80 & 72.40 & 72.77 \\

& & APoT~\cite{li2019additive}
& {}
& \best{16.37} & 65.57 & 70.61 & 72.15 & 72.54 \\

& & PowerQuant~\cite{yvinec2023powerquant}
& {}
& 15.94 & 66.03 & 71.14 & 72.41 & 72.70 \\

& & NANQ
& {}
& 15.28 & \best{66.58} & \best{71.61}
& \best{72.67} & \best{72.83} \\

\addlinespace[0.8pt]

& \multirow{4}{*}{Base}
& Uniform
& \multirow{4}{*}{85.11}
& 34.73 & 81.86 & 84.20 & 84.54 & 84.67 \\

& & APoT~\cite{li2019additive}
& {}
& 70.96 & 83.17 & 84.29 & 84.54 & 84.66 \\

& & PowerQuant~\cite{yvinec2023powerquant}
& {}
& 72.18 & 83.51 & 84.36 & 84.58 & \best{84.75} \\

& & NANQ
& {}
& \best{73.40} & \best{83.78} & \best{84.43}
& \best{84.61} & 84.71 \\

\bottomrule
\end{tabular*}

\vspace{4pt}

\textbf{(b) Language Models on WikiText-2: PPL $\downarrow$}

\vspace{1.5pt}

\begin{tabular*}{\textwidth}{
    @{\extracolsep{\fill}}
    lll*{6}{c}
    @{}
}
\toprule
Family
& Model
& Method
& Baseline
& $B_m=2$
& $B_m=3$
& $B_m=4$
& $B_m=5$
& $B_m=6$ \\
\midrule

\multirow{8}{*}{Pythia}
& \multirow{4}{*}{410M}
& Uniform
& \multirow{4}{*}{16.96}
& 7.61k & 46.04 & 24.87 & 22.10 & 21.70 \\

& & APoT~\cite{li2019additive}
& {}
& 615.3 & 32.13 & 24.00 & 22.32 & 21.92 \\

& & PowerQuant~\cite{yvinec2023powerquant}
& {}
& 431.8 & 30.82 & 23.55 & 22.05 & 22.08 \\

& & NANQ
& {}
& \best{294.1} & \best{29.73} & \best{23.10}
& \best{21.84} & \best{21.64} \\

\addlinespace[0.8pt]

& \multirow{4}{*}{1B}
& Uniform
& \multirow{4}{*}{11.61}
& 2.54k & 16.64 & 13.19 & 12.64 & 12.49 \\

& & APoT~\cite{li2019additive}
& {}
& 184.1 & 14.55 & 12.95 & 12.65 & 12.50 \\

& & PowerQuant~\cite{yvinec2023powerquant}
& {}
& 112.6 & 14.31 & 12.83 & 12.58 & \best{12.45} \\

& & NANQ
& {}
& \best{66.83} & \best{14.10} & \best{12.72}
& \best{12.53} & 12.47 \\

\midrule






\multirow{8}{*}{Llama-3.2}
& \multirow{4}{*}{1B}
& Uniform
& \multirow{4}{*}{8.620}
& 35.0k & 43.99 & 11.56 & 10.12 & 9.869 \\

& & APoT~\cite{li2019additive}
& {}
& 3.56k & 16.94 & 10.56 & 10.02 & 9.886 \\

& & PowerQuant~\cite{yvinec2023powerquant}
& {}
& 1.84k & 14.72 & 10.41 & 9.950 & 9.840 \\

& & NANQ
& {}
& \best{650.6} & \best{12.87} & \best{10.30}
& \best{9.888} & \best{9.808} \\

\addlinespace[0.8pt]

& \multirow{4}{*}{3B}
& Uniform
& \multirow{4}{*}{6.941}
& 11.1k & 13.62 & 8.223 & 7.619 & 7.495 \\

& & APoT~\cite{li2019additive}
& {}
& 916.1 & 9.916 & 7.777 & 7.566 & 7.511 \\

& & PowerQuant~\cite{yvinec2023powerquant}
& {}
& 482.7 & 9.210 & 7.720 & \best{7.480} & 7.489 \\

& & NANQ
& {}
& \best{129.1} & \best{8.557} & \best{7.668}
& 7.512 & \best{7.476} \\

\midrule

\multirow{4}{*}{OPT}
& \multirow{4}{*}{1.3B}
& Uniform
& \multirow{4}{*}{12.56}
& 13.4k & 286.6 & 14.92 & 13.81 & 13.79 \\

& & APoT~\cite{li2019additive}
& {}
& 7.62k & 23.51 & 14.04 & 13.85 & 13.77 \\

& & PowerQuant~\cite{yvinec2023powerquant}
& {}
& 4.18k & 19.07 & 14.18 & 13.80 & \best{13.75} \\

& & NANQ
& {}
& \best{1.54k} & \best{15.16} & \best{13.92}
& \best{13.77} & 13.76 \\

\bottomrule
\end{tabular*}

\caption{On-chip inference performance on (a) vision models using
CIFAR-100 and ImageNet-1K~\cite{imagenet}, and
(b) language models using WikiText-2. Quantized results are
averaged over three independent on-chip inference runs; Baseline denotes clean
BF16 inference. Shaded bold values indicate the best quantized result for each
model and $B_m$.}
\label{tab:onchip_models_noisy}

\endgroup
\vspace{-1.5mm}
\end{table*}

\subsubsection{End-to-End Inference Evaluation}

Many non-uniform quantization methods require retraining or iterative
optimization of thresholds and codebooks. We compare NANQ with Uniform
quantization, APoT, and PowerQuant, a representative data-free non-uniform
method. CIMQ and CIM$^2$PQ jointly optimize
array-wise input, weight, and partial-sum precision under architecture-specific
hardware objectives and are therefore not directly comparable under our
weight-only evaluation setting. $B_m$ denotes the weight-magnitude bit-width
excluding the sign bit and ranges from 2 to 6 bits, while activations are fixed
at INT8.

As shown in Table~\ref{tab:onchip_models_noisy}, NANQ achieves the best
mean performance in 38 of the 45 model--bit-width configurations, with the
largest gains at $2$--$3$ bits. Compared with PowerQuant, at $B_m=2$, NANQ
improves the accuracy of ResNet-20, VGG-11, and ViT-Base by $7.21$, $24.41$,
and $1.22$ percentage points, respectively, and reduces language-model PPL by
up to $73.3\%$. At $B_m=3$, its PPL reductions over PowerQuant range from
$1.5\%$ to $20.50\%$, with smaller model variants generally benefiting more,
suggesting that larger models are inherently more robust to low-bit
perturbations. The advantage narrows at higher precision, with PowerQuant
slightly outperforming NANQ in a few $5$--$6$-bit settings, as denser
quantization levels reduce the benefit of noise-aware level allocation.
Overall, NANQ is particularly effective for low-bit analog CIM inference where
hardware noise is dominant.

\begin{figure}[!t]
\centering
\includegraphics[width=1.0\columnwidth]{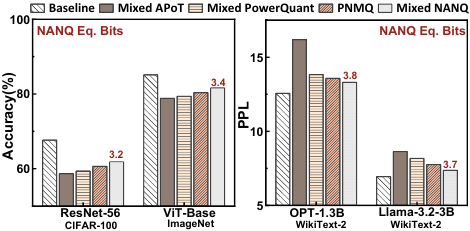}
\caption{Comparison of mixed-precision quantization methods on vision and language models ($\tau$ = 0.10).}
\label{fig:mixed}
\end{figure}

We further evaluate mixed-precision quantization under matched precision
budgets. For each model, NANQ uses $\tau=0.1$ to determine the layer-wise
bit-widths and the corresponding target equivalent bit-width
$B_{\mathrm{eq}}$. Under the same budget, APoT and PowerQuant start from the
minimum bit-width and greedily increase the precision of the layer with the
largest marginal gain per added weight bit until the budget is exhausted.
PNMQ~\cite{mixed} follows its original allocation strategy under the same
budget. This protocol allows each method to independently optimize its
layer-wise configuration while avoiding confounding effects from different
quantization resources.

As shown in Figure~\ref{fig:mixed}, Mixed NANQ achieves the best quantized
performance on all evaluated models, with adaptively determined equivalent
bit-widths ranging from $3.2$ to $3.8$ bits. Compared with PNMQ, NANQ improves
accuracy by $1.26$ percentage points on both CIFAR-100 ResNet-56 and
ImageNet-1K ViT-Base, while reducing the PPL of OPT-1.3B and Llama-3.2-3B from
$13.57$ and $7.760$ to $13.31$ and $7.361$, respectively. Compared with Mixed
PowerQuant, NANQ improves accuracy by $2.52$ and $2.23$ percentage points on
the two vision models and reduces PPL by $3.8\%$ and $9.9\%$ on the two
language models. These results show that NANQ improves performance under matched resources through noise-aware quantization and layer-wise precision allocation.

\newcommand{\ppl}[2]{%
  \begin{tabular}[c]{@{}c@{}}
    {\small #1}\\[-1.5pt]
    {\scriptsize $\pm$#2}
  \end{tabular}%
}

\newcommand{\bestppl}[2]{%
  \begin{tabular}[c]{@{}c@{}}
    {\small\bfseries #1}\\[-1.5pt]
    {\scriptsize\bfseries $\pm$#2}
  \end{tabular}%
}

\begin{table}[t]
\centering

\footnotesize
\renewcommand{\arraystretch}{1.12}
\setlength{\tabcolsep}{1.5pt}

\begin{tabular*}{\linewidth}{
  @{\extracolsep{\fill}}
  >{\small}c
  >{\small}c
  ccccc
  @{}
}
\toprule
\multirow{2}{*}{\small\boldmath{$B_m$}}
& \multirow{2}{*}{\small\boldmath{$\gamma^*$}}
& \multicolumn{5}{c}{$\gamma=\gamma^*+\Delta\gamma$} \\
\cmidrule(lr){3-7}
& & $-0.2$ & $-0.1$ & $0$ & $+0.1$ & $+0.2$ \\
\midrule

2 & 1.23
& \ppl{392.973}{46.334}
& \ppl{370.075}{20.973}
& \bestppl{294.091}{16.292}
& \ppl{303.869}{15.394}
& \ppl{338.122}{36.829} \\

3 & 1.15
& \ppl{32.588}{0.254}
& \ppl{31.014}{0.318}
& \bestppl{29.728}{0.286}
& \ppl{30.314}{0.254}
& \ppl{32.433}{0.604} \\

4 & 1.09
& \ppl{23.206}{0.173}
& \ppl{23.675}{0.180}
& \bestppl{23.102}{0.206}
& \ppl{23.177}{0.310}
& \ppl{23.476}{0.352} \\

5 & 0.93
& \ppl{22.078}{0.124}
& \ppl{21.900}{0.216}
& \bestppl{21.842}{0.153}
& \ppl{22.004}{0.277}
& \ppl{21.991}{0.249} \\

6 & 1.42
& \ppl{21.749}{0.188}
& \ppl{21.712}{0.173}
& \bestppl{21.642}{0.193}
& \ppl{21.717}{0.242}
& \ppl{21.732}{0.196} \\

\bottomrule
\end{tabular*}

\caption{PPL of Pythia-410M on WikiText-2 under perturbations around
the selected $\gamma^*$. Results are reported as mean $\pm$ standard
deviation. Lower is better.}
\label{tab:gamma_ablation}

\end{table}

\begin{table}[t]
\centering
\small
\renewcommand{\arraystretch}{1.08}
\setlength{\tabcolsep}{13pt}
\begin{tabular}{ccc}
\toprule
$\tau$
& $B_{\mathrm{eq}}$
& PPL $\downarrow$ \\
\midrule
0
& 5.86
& $21.87 \pm 0.18$ \\

0.05
& 4.63
& $22.04 \pm 0.20$ \\

\rowcolor{gray!15}
\textbf{0.10}
& \textbf{3.76}
& $\mathbf{22.68 \pm 0.22}$ \\

0.20
& 3.42
& $24.74 \pm 0.27$ \\

0.30
& 3.16
& $26.31 \pm 0.34$ \\

0.50
& 2.72
& $29.68 \pm 0.48$ \\
\bottomrule
\end{tabular}
\caption{Effect of $\tau$ on the equivalent bit-width and PPL of
Pythia-410M on WikiText-2. Results are reported as mean $\pm$ standard
deviation.}
\label{tab:tau_ablation}
\end{table}

\subsection{Ablation Study}
Table~\ref{tab:gamma_ablation} evaluates the sensitivity to the selected
$\gamma^*$ by perturbing it within
$\pm 0.2$. The selected $\gamma^*$ consistently achieves the lowest mean
PPL for all $B_m$ values, validating our parameter selection. The impact
of $\gamma$ is most significant at $B_m=2$, where the PPL increases from
$294.091$ to as high as $392.973$ and the standard deviation also becomes
larger. As $B_m$ increases, the maximum PPL degradation decreases from
about $9.6\%$ at $B_m=3$ to below $0.5\%$ at $B_m=6$, indicating that
$\gamma$ is critical under aggressive low-bit quantization, while the
method remains robust at higher bit-widths.

Table~\ref{tab:tau_ablation} shows the effect of the saturation threshold
$\tau$ on the equivalent bit-width and model performance. Decreasing $\tau$
from $0.1$ to $0.05$ and $0$ increases $B_{\mathrm{eq}}$ from $3.76$ to
$4.63$ and $5.86$, while improving PPL only modestly from $22.68$ to
$22.04$ and $21.87$. In contrast, increasing $\tau$ to $0.2$ or above
further reduces the equivalent bit-width but noticeably increases both PPL
and its standard deviation. These results indicate that $\tau=0.1$ provides
a favorable trade-off between model performance and equivalent precision.

\section{Discussion and Limitations}

NANQ can be extended to other multi-level analog memories through
device-specific noise characterization, although substantial changes in
temperature, aging, or operating conditions may require the noise profile to be
updated. The reported $B_{\mathrm{eq}}$ is an algorithm-level precision metric
rather than a direct measure of physical storage, latency, or energy savings;
realizing such benefits requires hardware-aware mapping and compiler support.
In addition, the current framework approximates CIM non-idealities as
magnitude-dependent weight perturbations. Future work may separately model and
jointly optimize quantization for weights, activations, and outputs.

\section{Conclusion}

This work presents NANQ, a training-free, noise-aware mixed-precision
non-uniform quantization framework for analog CIM. NANQ adaptively assigns
quantization levels according to chip-measured magnitude-dependent weight
noise and determines layer-wise bit-widths from their precision saturation
under hardware noise. On-chip experiments on an eFlash CIM SoC show that NANQ
significantly outperforms existing methods at low precision, while
mixed-precision NANQ captures most attainable quantization gains using only
$3.2$--$3.8$ equivalent bits. These results demonstrate that explicitly
accounting for the hardware noise floor enables more efficient use of limited
precision in analog CIM inference.


\clearpage
\bibliography{aaai2027}


\end{document}